\pdfoutput=1

\documentclass[11pt]{article}

\usepackage[]{ACL2023}

\usepackage{times}
\usepackage{latexsym}

\usepackage[T1]{fontenc}

\usepackage[utf8]{inputenc}

\usepackage{microtype}

\usepackage{inconsolata}

\usepackage{natbib} 
\usepackage{adjustbox}
\usepackage{graphicx}
\usepackage{algorithm}
\usepackage{algorithmic}
\usepackage{subcaption}
\usepackage{amsmath}
\usepackage{bm}
\usepackage{multirow}
\usepackage{xcolor}
\usepackage[utf8]{inputenc}
\usepackage{mathtools}
\usepackage{siunitx}
\usepackage{float}
\usepackage{amsfonts}
\usepackage[T1]{fontenc}
\usepackage[utf8]{inputenc}

\title{Asymmetric feature interaction for interpreting model predictions}

\author{
Xiaolei Lu$^1$ \quad \bf Jianghong Ma  $^2$ \quad  \bf Haode Zhang$^3$ \\ 
Department of Electrical Engineering, City University of Hong Kong, Hong Kong S.A.R.$^1$\\
School of Computer Science and Technology,Harbin Institute of Technology,Shenzhen,China$^2$ \\
Department of Computing, The Hong Kong Polytechnic University, Hong Kong S.A.R.$^3$ \\
}

\usepackage{hyperref}

\begin{document}

\maketitle

\begin{abstract}
In natural language processing (NLP), deep neural networks (DNNs) could model complex interactions between context and have achieved impressive results on a range of NLP tasks. Prior works on feature interaction attribution mainly focus on studying symmetric interaction that only explains the additional influence of a set of words in combination, which fails to capture asymmetric influence that contributes to model prediction. In this work, we propose an asymmetric feature interaction attribution explanation model that aims to explore asymmetric higher-order feature interactions in the inference of deep neural NLP models. By representing our explanation with an directed interaction graph, we experimentally demonstrate interpretability of the graph to discover asymmetric feature interactions. Experimental results on two sentiment classification datasets show the superiority of our model against the state-of-the-art feature interaction attribution methods in identifying influential features for model predictions.

\end{abstract}

\section{Introduction}

Deep neural networks (DNNs) have demonstrated impressive results on a range of Natural Language Processing (NLP) tasks. Unlike traditional models (e.g. CRFs and HMMs) that optimize weights on human interpretable features, deep neural models operate like a black box by applying multiple layers of non-linear transformation on the vector representations of text data, which fails to provide insights to understand the inference process of deep neural models over the features (e.g. words and phrases) involved in modeling.

\begin{figure}[H]
\centering
\includegraphics[width=1\linewidth,height = 0.8in ]{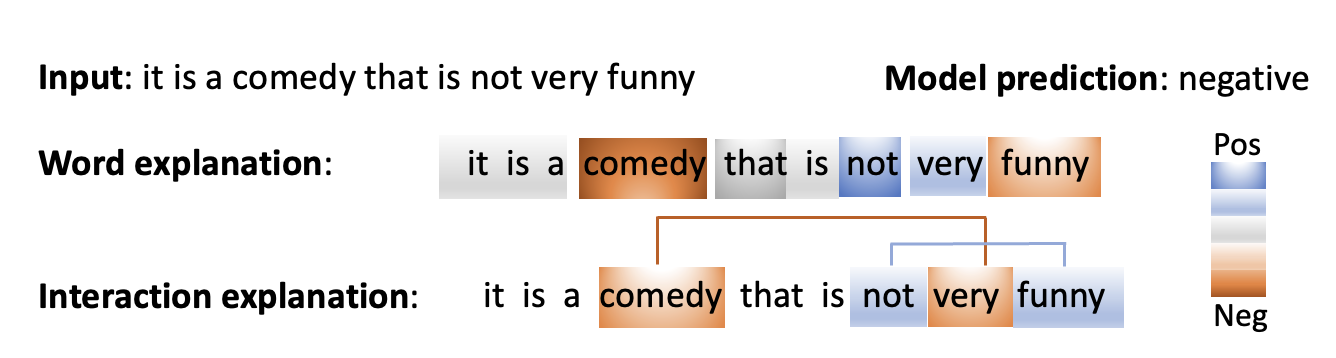}
\caption {Explanations for a negative
movie review (computed by Shapley value and Shapley interaction index), where the color indicates contribution of the corresponding word/pairwise word interaction to the model prediction.} 
\label{fig1}
\end{figure}

Interpreting the prediction of a black box model could help understand model inference behaviors and increase user trust in applying the model to real-world applications. Prior efforts in NLP mainly focus on quantifying the contributions of individual word or word interactions to the prediction. Fig.\ref{fig1} demonstrates word-level and pairwise word interaction explanations for a sentiment classification task, where the word ``not'' and the interaction between ``not'' and ``funny'' contribute positively to the prediction \textbf{Negative}.

\begin{figure}[H]
\centering
\includegraphics[width=0.95\linewidth,height = 0.9in ]{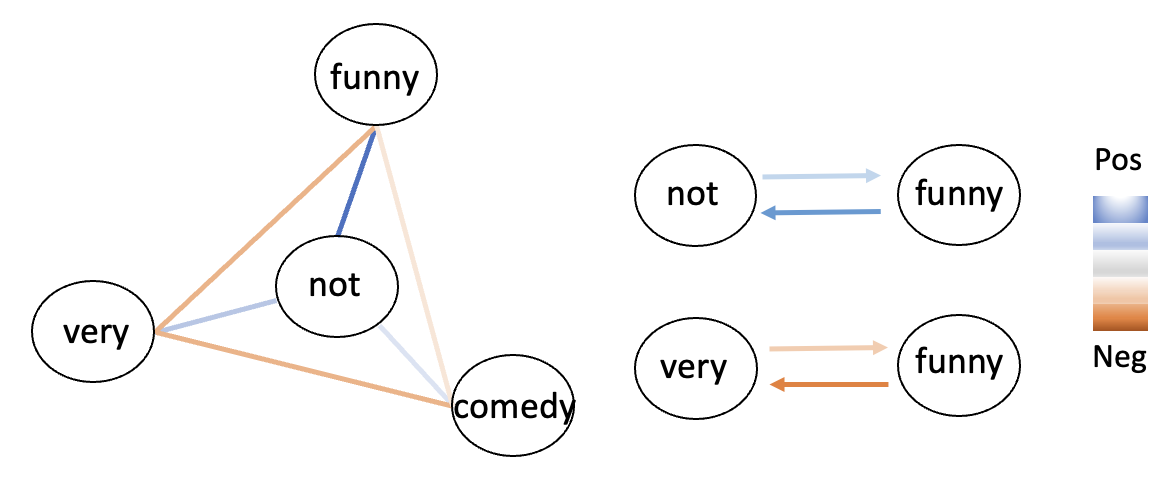}
\caption {Symmetric versus asymmetric pairwise interaction (computed by our method) where the directed edge $a\rightarrow b$ refers to in the presence of $a$ how much contribution of $b$ made to the model prediction.  The presence of ``very'' does not influence ``funny'' much while ``funny'' further modifies ``very'' and thus the interaction influence of $``\mathrm{funny}"\rightarrow ``\mathrm{very}" $ is stronger than that of $``\mathrm{very}"\rightarrow``\mathrm{funny}" $.
} 
\label{fig2}
\end{figure}
Studying word interaction could help identify to what extent a set of words exert influence in combination as opposed to independently. However, most interaction attribution methods assume symmetric interaction, which may fail to capture asymmetric influence that contributes to model prediction. Fig.~\ref{fig2} presents some symmetric pairwise interactions with graph representation for the instance in Fig.~\ref{fig1}, where words becomes nodes and edges between words represent interaction. In individual-level explanation ``funny'' has negative influence while the symmetric interaction between ``funny" and ``not" produces positive influence to model prediction. Therefore the influence of the presence of ``not'' to ``funny'' is not the same as that of the presence of ``funny'' to ``not". ``funny'' has weak positive contribution in the presence of ``not'' and ``funny'' further modifies ``not'', the interaction influence of $``\mathrm{funny}"\rightarrow ``\mathrm{not}" $ could be stronger than the interaction of $``\mathrm{not}"\rightarrow``\mathrm{funny}"$. For the ideal asymmetric interaction, the presence of other features should not negate the positive influence of the important feature and other features would lose their influence when important features are present. Constructing asymmetric interaction graph for the predicted instance could help human have a nuanced understanding toward the inference of deep NLP models.

In this paper, our work aims to provide the explanation that incorporates asymmetric feature interaction \footnote{Our code is available at \url{https://github.com/StillLu/ASIV}.}. The contributions are summarized as follows:

\begin{itemize}

\item We propose an asymmetric feature interaction attribution method that incorporates asymmetric higher-order feature interactions toward explaining the prediction of deep neural NLP models. 

\item We investigate three different sampling strategies in NLP field for computing marginal contribution of our defined asymmetric feature interaction attribution score, and empirically show that none is generally better than the others or more broadly applicable.

\item We evaluate the proposed model on two sentiment classification datasets with BERT \citep{devlin2018bert} and RoBERTa \citep{liu2019roberta}, and the experimental results demonstrate the faithfulness of our explanation model.
\end{itemize}

\section{Related work}

\subsection{Feature attribution explanation}
Most explanation methods mainly focus on model-agnostic explanations and study how to effectively measure the importance of features on the prediction. For example, LIME \citep{zhang2019should} evaluates the contribution of each feature by learning a linear model locally around a instance. Shapley value \citep{shapley1997value,lundberg2017unified} estimates the influence of a feature by averaging its marginal contribution among all permutations. Since the computation of Shapley value is computationally expensive, popular variants like Kernel SHAP \citep{lundberg2017unified} and  Quasi-random and adaptive sampling \citep{vstrumbelj2014explaining} are proposed to efficiently approximate Shapley value. However, these methods do not explain how feature interactions contribute to model predictions, which fails to address model's learning capability from high-order feature interactions.

\subsection{Feature interaction explanation}
There are increasingly research on studying feature interaction explanation methods. For example, Shapley interaction index \citep{grabisch1997k} and Shapley Taylor interaction index \citep{dhamdhere2019shapley} are proposed to measure the interaction between multiple players. Integrated Directional Gradients (IDG) \citep{sikdar2021integrated} borrowed axioms from Integrated Gradients (IG) \citep{sundararajan2017axiomatic}, where the desirable characteristics are satisfied by IG and Shapley value in cooperative game theory. \citet{tsang2020does} proposed an efficient framework Archipelago to combine feature interaction detector (ArchDetect) and feature attribution measure (ArchAttribute). 

Further, feature interaction could be explained in a hierarchical structure. Agglomerative contextual decomposition (ACD) \citep{singh2018hierarchical} builds the hierarchical explanations in a bottom-up way by starting with individual features and iteratively combining them based on the generalized CD scores. \citet{jin2019towards} addressed context independent importance that is ignored in ACD, and proposed an easy and model-agnostic Sampling and Occlusion (SOC) algorithm to incorporate conditional context information given the specified text sequence. HEDGE \citep{chen2020generating} designs a top-down framework to construct hierarchical explanations by detecting the weakest interaction point and selecting important sub-span for a given text span. 

The above feature interaction explanation methods only focus on symmetric interaction. We note that concurrent work by \citet{masoomi2021explanations} addressed directed pairwise interaction. However, their Shapley value based formulation also introduces noisy interaction as different subsets may contain several same elements. Moreover, this solution ignores asymmetric high-order interaction.

\section{Asymmetric Shapley interaction value}

This section first revisits Shapley value for explaining model prediction, then describes the definition of asymmetric Shapley interaction value and the corresponding approximating computation.

The following notations will be used throughout the paper. For a classification task, given a text sequence  $\bm{x}=\left ( x_1,...,x_n \right )$ and a trained model $f$, $\hat{y}$ is the prediction label and $f(\cdot )$ denotes the model output probability on $\hat{y}$.

\subsection{Shapley value for model interpretability}

In cooperative game theory Shapley value measures the marginal contribution that a player makes upon joining the group by averaging over all possible permutations of players in the group.

The Shapley value of $i_{th}$ word in $\bm{x}$ for the model prediction $\hat{y}$ is weighted and summed over all possible word combinations:

\begin{equation}
\scalebox{0.95}{$
\phi(i)=\sum_{S\subseteq N\setminus {i}}\frac{(n-1-\left | S \right |)!\left | S \right |!}{n!}\left [ v(S\cup {i}) -v(S)\right ],$}
\end{equation}
where $S$ is the subset of feature indices. The equivalent formulation is 
\begin{equation}
\phi(i)=\frac{1}{n!}\sum\limits_{\mathcal{O}\in \pi(n)} v(\mathrm{pre}^i(\mathcal{O})\cup i) -v(\mathrm{pre}^i(\mathcal{O})),
\end{equation} 
where $\pi(n)$ denotes all permutation of the word indexes $\left \{ 1,2,...,n \right \}$. $\mathrm{pre}^i(\mathcal{O})$ is the set of all indices that precede $i$ in the permutation $\mathcal{O} \in \pi(n)$.

$v(S)$ is the value function that characterizes the contribution of the subset $S$ to the prediction $\hat{y}$:
\begin{equation}
v(S) =\mathbb{E}(f|\bm{x}_S\cup \bm{x}'_{\bar{S}}) -\mathbb{E}(f|\bm{x}'),
\end{equation}
where $\bm{x}'$ denotes the text sequence with the same length as $\bm{x}$, $ \bar{S}= N\setminus {S}$. $\mathbb{E}(f|\bm{x}_S\cup \bm{x}'_{\bar{S}})$ is the expectation of $f(\cdot )$ over possible $\bm{x}'$ where only the subset values $\bm{x}_S$ unchanged.

\subsection{Definition of Asymmetric Shapley interaction value}

The Shapley value above can quantify the contribution of a single word or phrase to the model prediction. The proposed asymmetric Shapley interaction value (ASIV) measures the asymmetric interaction between two different subsets $T_1$ and $T_2$ that attributes to the model prediction. That is, ASIV determines the contribution of $T_1$ conditioned on the presence of $T_2$ to the prediction $\hat{y}$. By treating $T_1$ and $T_2$ as two singletons among the players, ASIV is defined as

\begin{equation}
\scalebox{0.9}{$
\phi_{T_2}(T_1)=\sum\limits_{T_2\subseteq S\subseteq N\setminus {\left \{ T_1\right \}}}C_1\Delta_{T_1} v(S)-\Delta_{T_1}v(S\backslash T_2),$}
\end{equation}
where $C_1 =\frac{(n-\left | T_1\right |-\left | T_2 \right |+1-\left | S \right |)!\left | S \right |!}{(n-\left |T_1\right |-\left |T_2\right |+2)!}$. $\Delta_{T_2} v(S)$ and $\Delta_{T_1}(S)$ are given as
\begin{equation}
\Delta_{T_1} v(S) = v(S\cup T_1)-v(S),
\end{equation}
\begin{equation}
\Delta_{T_1} v(S\backslash T_2) = v(S\backslash T_2\cup T_1)-v(S\backslash T_2).
\end{equation}

$\Delta_{T_1} v(S)-\Delta_{T_1}v(S\backslash T_2)$ computes the difference of marginal contribution of $T_1$ to the coalition $S$ with and without participation of the subset $T_2$, which aims to capture directional interaction influence between $T_1$ and $T_2$. The equivalent formulation is 
\begin{equation}
\scalebox{0.8}{$
\phi_{T_2}{(T_1)}=C_2 \sum\limits_{\mathcal{O}} \Delta_{T_1} v(\mathrm{pre}^{T_1}(\mathcal{O}))-\Delta_{T_1}v(\mathrm{pre}_{T_2}^{T_1}(\mathcal{O})),$}
\end{equation}
where $\mathcal{O}$ denotes the possible permutation where $T_2$ precedes $T_1$, and $C_2=\frac{1}{(n-\left |T_1\right|-\left |T_2\right|+2)!/2}$. $\Delta_{T_1} v(\mathrm{pre}^{T_1}(\mathcal{O}))$ and $\Delta_{T_1}v(\mathrm{pre}_{T_2}^{T_1}(\mathcal{O}))$ are defined as
\begin{equation}
\scalebox{0.9}{$
\Delta_{T_1} v(\mathrm{pre}^{T_1}(\mathcal{O})) =  v(\mathrm{pre}^{T_1}(\mathcal{O})\cup T_1)-v(\mathrm{pre}^{T_1}(\mathcal{O})),$}
\end{equation}
\begin{equation}
\scalebox{0.9}{$
\Delta_{T_1}v(\mathrm{pre}_{T_2}^{T_1}(\mathcal{O})) = v(\mathrm{pre}_{T_2}^{T_1}(\mathcal{O})\cup T_1)-v(\mathrm{pre}_{T_2}^{T_1}(\mathcal{O})),$}
\end{equation}
where $\mathrm{pre}^{T_1}$ denotes the set of all indices that precede $T_1$ while $\mathrm{pre}^{T_1}_{T_2}$ excludes $T_2$ from this set.

If both $T_1$ and $T_2$ contain a single element (i.e. $T_1 = \left \{ i \right \}$ and $T_2 = \left \{ j \right \}$), the directed pairwise relationship could be obtained by

\begin{equation}
\begin{aligned}
\phi_{j}(i)=&\sum_{j\subseteq S\subseteq N\setminus {\left \{ i\right \}}} C_3 v(S\cup \left \{ i \right \})-v(S) \\
&\scalebox{0.9}{$-\bigl ( v(S\backslash \left \{ j \right \}\cup \left \{ i \right \})-v(S\backslash \left \{ j \right \})  \bigr )$},
\end{aligned} 
\end{equation}
where $C_3=\frac{(n-1-\left | S \right |)!\left | S \right |!}{n!}$.

\subsection{Approximating computation}

Computing asymmetric Shapley interaction value has to estimate value function $v(S)$ over all possible permutations (or subsets). First, we investigate three different sampling strategies of computing $\mathbb{E}(f|\bm{x}_S\cup \bm{x}'_{\bar{S}})$. 

\textbf{Marginal Expectation (ME)}: In applying Shapley value to explain predictions of NLP models, prior research assumes individual features are mutually independent. Then $\mathbb{E}(f|\bm{x}_S\cup \bm{x}'_{\bar{S}})$ is computed as
\begin{equation}
\mathbb{E}(f|\bm{x}_S\cup \bm{x}'_{\bar{S}})=\mathbb{E}_{p(\bm{x}'_{\bar{S}})}(f|\bm{x}_S\cup \bm{x}'_{\bar{S}}),
\end{equation}
where $\bm{x}'_{\bar{S}}$ could be randomly sampled from training data or a sequence with all $\left \langle \mathrm{pad} \right \rangle$ tokens. 

In computing $f(\bm{x}_S\cup \bm{x}'_{\bar{S}})$, the combination of $\left \{ \bm{x}_S\cup \bm{x}'_{\bar{S}} \right \}$ may be incompatible. For example, given the instance $\bm{x}$ ``\emph{the issue of faith is not explored deeply}'' and the subset $\bm{x}_S$ ``\emph{the issue of faith}'', random sampling could generate the sequence ``\emph{the issue of faith \textbf{time changer may not}}''. Such incoherence lie \textbf{off the data manifold} \citep{frye2020shapley} where $v(S)$ may fail to capture model's dependence on the whole context information.

\textbf{Conditional Expectation (CE)}: The expectation of $(f|\bm{x}_S\cup \bm{x}'_{\bar{S}})$ with respect to the distribution $\bm{x}'_{\bar{S}}$ conditioning on $\bm{x}_S$ is computed as 
\begin{equation}
\mathbb{E}(f|\bm{x}_S\cup \bm{x}'_{\bar{S}})=\mathbb{E}_{p(\bm{x}'_{\bar{S}}|\bm{x}_S)}(f|\bm{x}_S\cup \bm{x}'_{\bar{S}}),
\end{equation}
where $\bm{x}'_{\bar{S}}|\bm{x}_S$ could be sampled from a pre-trained language model.

In conditional expectation, the combined input $\bm{x}_S\cup \bm{x}'_{\bar{S}}$ is more coherent. For example, given the subset $\bm{x}_S$ ``\emph{the issue of faith}'', the generated complete sequence could be ``\emph{the issue of faith \textbf{is not very important}}''.

However, both marginal expectation $\mathbb{E}_{p(\bm{x}'_{\bar{S}})}(f|\bm{x}_S\cup \bm{x}'_{\bar{S}})$ and conditional expectation $\mathbb{E}_{p(\bm{x}'_{\bar{S}}|\bm{x}_S)}(f|\bm{x}_S\cup \bm{x}'_{\bar{S}})$ have to evaluate $f$ on some \textbf{out-of-domain} data, where the model $f$ is forced to extrapolate to an unseen part of the feature space. \citet{hooker2021unrestricted} conducted many simulation experiments showing that the permutation-based feature attributions are sensitive to these edge cases.

\textbf{In-domain Expectation (IDE)}: To enable $f$ to evaluate on \textbf{in-domain} data, we pretrain a language model on the training data $\bm{\mathrm{x}}=\left \{\bm{x}_1,...,\bm{x}_N \right \}$ to model the underlying data distribution $p(\bm{\mathrm{x}})$. For a text sequence $\bm{x}'$ with known subset $\bm{x}'_S =\bm{x}_S $, by sampling the remaining subset $\bm{x}'_{\bar{S}}$ from $p(\bm{\mathrm{x}})$, $\mathbb{E}(f|\bm{x}_S\cup \bm{x}'_{\bar{S}})$ is computed as
\begin{equation}
\mathbb{E}(f|\bm{x}_S\cup \bm{x}'_{\bar{S}})=\mathbb{E}_{\bm{x}'_{\bar{S}}|\bm{x}_S\sim p(\bm{\mathrm{x}})}(f|\bm{x}_S\cup \bm{x}'_{\bar{S}}).
\end{equation}

Second, by enumerating all possible permutations, Eq.(7) could be replaced with 

\begin{equation}
\scalebox{0.8}{$
\begin{aligned}
\phi_{T_2}{(T_1)}= C_2 \sum\limits_{\mathcal{O}} \sum_{\bm{w}} &p(\bm{w})f(\bm{w}_{[\bm{w}_j=\bm{x}_j,j\in\mathrm{pre}^{T_1}(\mathcal{O})\cup T_1]}) \\
&-f(\bm{w}_{[\bm{w}_j=\bm{x}_j,j\in\mathrm{pre}^{T_1}(\mathcal{O})]})\\
&-\large\left [ f(\bm{w}_{[\bm{w}_j=\bm{x}_j,j\in\mathrm{pre}_{T_2}^{T_1}(\mathcal{O})\cup T_1]})\right.\\
& \left.-f(\bm{w}_{[\bm{w}_j=\bm{x}_j,j\in\mathrm{pre}_{T_2}^{T_1}(\mathcal{O})]})\large\right],
\end{aligned}
$}
\end{equation}
where $\bm{w}$ could be obtained from the above three different sampling strategies.

Let $ V_{\mathcal{O},w}$ for all permutation/instance pairs, 
\begin{equation}
\scalebox{0.75}{$
\begin{aligned}
 V_{\mathcal{O},w} = & f(\bm{w}_{[\bm{w}_j=\bm{x}_j,j\in\mathrm{pre}^{T_1}(\mathcal{O})\cup T_1]}) -f(\bm{w}_{[\bm{w}_j=\bm{x}_j,j\in\mathrm{pre}^{T_1}(\mathcal{O})]})\\
&-\large\left [ f(\bm{w}_{[\bm{w}_j=\bm{x}_j,j\in\mathrm{pre}_{T_2}^{T_1}(\mathcal{O})\cup T_1]})-f(\bm{w}_{[\bm{w}_j=\bm{x}_j,j\in\mathrm{pre}_{T_2}^{T_1}(\mathcal{O})]})\large\right],
 \end{aligned}
 $}
\end{equation}
we adopt Monte Carlo sampling \citep{vstrumbelj2014explaining} to approximate $ \phi_{T_2}{(T_1)} $ as
\begin{equation}
  \phi_{T_2}{(T_1)} = \frac{1}{m}\sum_{j=1}^m V_j.
\end{equation}

\section{Experiments}
We evaluate explanation methods on text classification tasks with BERT \citep{devlin2018bert} and RoBERTa \citep{liu2019roberta} models. 

\subsection{Datasets and classification models}
We use Stanford Sentiment Treebank-2 (SST-2) dataset with $6920/1821$ in train/test sets \citep{socher2013recursive} and the Yelp Sentiment Polarity (Yelp-2) dataset with $560,000/38,000$ in train/test sets \citep{zhang2015character}. The average review length in SST-2 test set is $19.25$ and $136.49$ in Yelp-2 test set. Also, we employ pretrained BERT-base and RoBERTa-base models, and then fine-tune them in the downstream classification tasks. BERT achieves accuracy $91.15\%$ on SST-2 and $96.39\%$ on Yelp-2, and RoBERTa achieves accuracy $94.39\%$ on SST-2 and $96.94\%$ on Yelp-2.

\subsection{Baselines}
We compare our method against five model-agnostic feature interaction attribution methods: HEDGE \citep{chen2020generating}, SOC \citep{jin2019towards}, Archipelago \citep{tsang2020does}, Shapley interaction index \citep{grabisch1997k} and Bivariate Shapley value \citep{masoomi2021explanations}. More implementation details of the baselines and ASIV could be found in Appendix \ref{appendixa}.

\subsection{Evaluation}

For asymmetric interaction between features, the important features are expected to have more positive incoming edges and less outgoing edges. Therefore to evaluate faithfulness of feature interaction attribution methods in deriving feature interaction relationship, in this paper we focus on pairwise interaction and apply PageRank \citep{page1999pagerank} algorithm to obtain feature importance ranking. Two evaluation metrics \citep{chen2020generating,nguyen2018comparing,shrikumar2017learning} are employed to evaluate influential features as follows: 

The area over the perturbation curve (AOPC): average change in the model output probability on the predicted class among the test data by deleting top $k$ influential words from each text sequence.

\begin{equation}
    \mathrm{AOPC}(k) = \frac{1}{N}\sum_{i=1}^N f(\bm{x}_i)-f(\tilde{\bm{x}}_i),
\end{equation}
where $\tilde{\bm{x}}_i$ is obtained by dropping the $k\%$ top-scored words from $\bm{x}_i$. The higher AOPC, the more important deleted words for model prediction.

Log-odds (LOR): average the difference of negative logarithmic probabilities on the predicted class over the test
data before and after replacing the top $k\%$ influential words with ``<pad>" token in the text sequence.

\begin{equation}
    \mathrm{LOR}(k) = \frac{1}{N}\sum_{i=1}^N \log \frac{f(\bm{x'}_i)}{f(\bm{x}_i)},
\end{equation}
where $\bm{x'}_i$ is obtained by masking the $k\%$ top-scored words from $\bm{x}_i$. The lower LOR, the more important deleted words for model prediction.

\subsection{Qualitative Analysis}

We first demonstrate interpretability of the feature asymmetric interaction graph by presenting a test example ``\emph{you might not buy the ideas}'' from SST-2 that is predicted as \textbf{Negative} by BERT model. More examples are shown in Appendix \ref{appendixc}. 

Fig.~\ref{fig11} focuses on two words ``not'' and ``might'' that are more consistent with human explanation and describes three feature interaction graphs estimated by the typical feature interaction attribution methods. Bivariate Shapley value also estimates asymmetric interaction relationship and Shapley interaction index models symmetric feature interaction. Here ASIV is estimated with random sampling as it performs best in SST-2 dataset. In the directed weighted interaction graph, as shown in Fig.~\ref{fig11} (a) and (b), $\mathrm{word}_1\overset{0.05}{\rightarrow}\mathrm{word}_2$ denotes that in the presence of $\mathrm{word}_1$, the influence score of $\mathrm{word}_2$ to model prediction is $0.05$. Fig.~\ref{fig11} (c) shows an undirected weighted interaction graph where the edge weight is the symmetric interaction influence to model prediction.

\begin{figure}[H]
\centering
\begin{subfigure}[b]{0.55\textwidth}
   \includegraphics[width=0.85\linewidth,height=5.5cm]{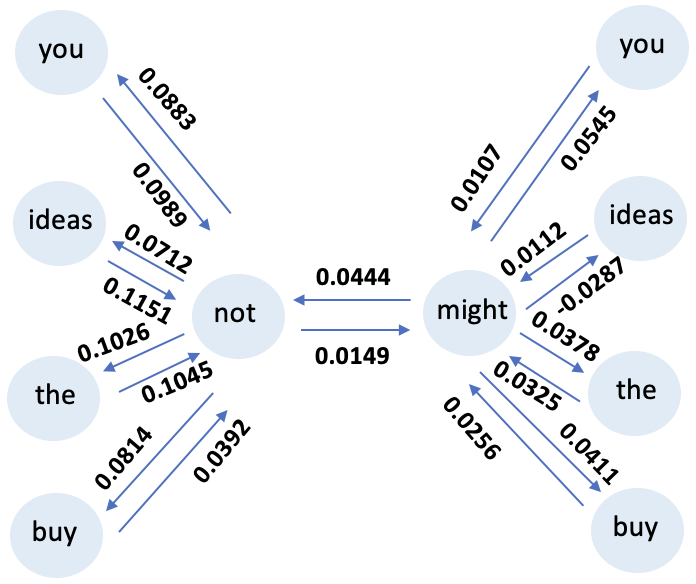}
   \caption{ASIV estimation.}
   \label{fig:Ng1} 
\end{subfigure}

\begin{subfigure}[b]{0.55\textwidth}
   \includegraphics[width=0.85\linewidth,height=5.5cm]{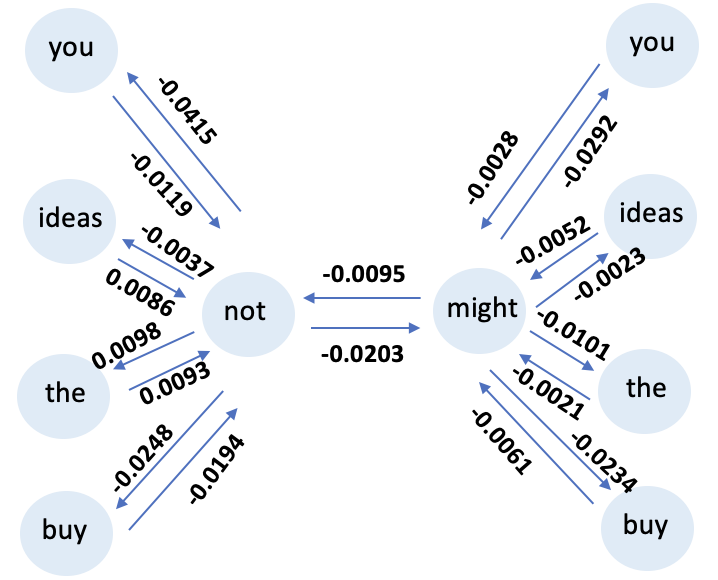}
   \caption{Bivariate Shapley estimation.}
   \label{fig:Ng2}

\end{subfigure}

 \begin{subfigure}[b]{0.55\textwidth}
   \includegraphics[width=0.85\linewidth,height=5.5cm]{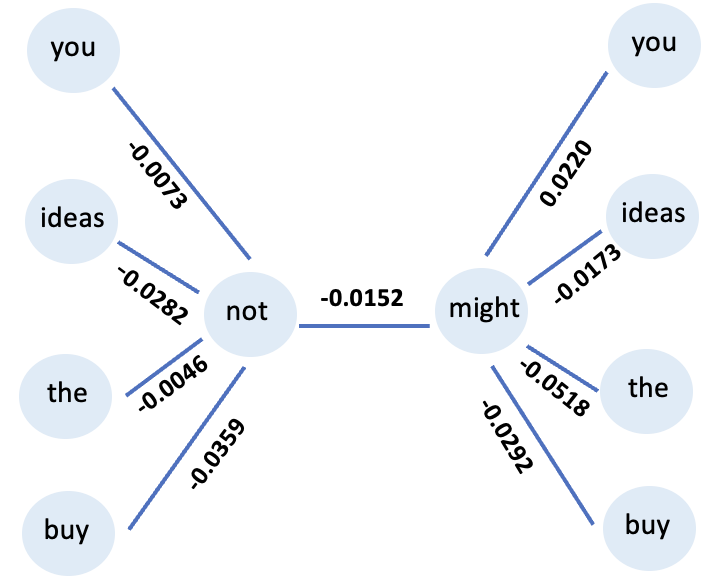}
   \caption{Shapley interaction index estimation.}
   \label{fig:Ng2}  
\end{subfigure}

\caption{Visualization of feature interaction graph estimated by ASIV, Bivariate Shapley value and Shapley interaction index.}

\label{fig11}
\end{figure}

\renewcommand{\arraystretch}{1.5}
\begin{table*}[]

\centering
\scalebox{0.9}{
\begin{tabular}{llllllllll} 
\hline
\multicolumn{2}{l}{\multirow{3}{*}{Models}}     & \multicolumn{4}{c}{SST-2}                                                                               & \multicolumn{4}{c}{Yelp-2}                                                                               \\ 
\cline{3-10}
\multicolumn{2}{l}{}                            & \multicolumn{2}{c}{~BERT}                          & \multicolumn{2}{c}{RoBERTa}                        & \multicolumn{2}{c}{BERT}                           & \multicolumn{2}{c}{RoBERTa}                         \\
\multicolumn{2}{l}{}                            & \multicolumn{1}{c}{AOPC} & \multicolumn{1}{c}{LOR} & \multicolumn{1}{c}{AOPC} & \multicolumn{1}{c}{LOR} & \multicolumn{1}{c}{AOPC} & \multicolumn{1}{c}{LOR} & \multicolumn{1}{c}{AOPC} & \multicolumn{1}{c}{LOR}  \\ 
\hline
\multicolumn{2}{l}{Hedge}                       &   0.0651                      &  -0.1488                       &  0.1262                       &  -0.2796                       &  0.0459                        &  -0.1061                       &   0.0218                       & -0.0337                         \\
\multicolumn{2}{l}{SOC}                         &     0.0761                   & - 0.1977                       &  0.0682                        &  -0.1473                       &  0.0363                        &  -0.0941                     &      0.0549                    &   -0.1545                       \\
\multicolumn{2}{l}{Archipelago}                 &    0.1055                      &  -0.2739                       & 0.0808                         &  -0.2292                       &  0.0461                        &  -0.0981                       &  0.0356                        &   -0.1591                       \\
\multicolumn{2}{l}{Bivariate Shapley~}          &    0.0787                     & -0.1958                        &  0.0916                        &   -0.2338                      &   0.0356                       &  -0.1254                       &  0.0162                        &  -0.0435                        \\
\multicolumn{2}{l}{Shapley interaction index}   &    0.0932                     &  -0.2268                       &  0.0845                        &  -0.2010                       & 0.0210                         &   -0.0403                      &  0.0266                        &  -0.0857                        \\
\multirow{4}{*}{Ours} & ME-random               &   \textbf{0.1929}                      &   \textbf{-0.5427}                      & \textbf{0.4286}                         & \textbf{-1.4086}                        &  0.1151                        &   -0.3149                      &   0.1026                       & -0.2971                         \\
                      & ME-padding              &  0.1484                        &  -0.3947                       &  0.1659                        &  -0.4832                       &  0.1109                        &  \textbf{-0.3755}                       &  0.0608                        & -0.2198                         \\
                      & CE                      &   0.1558                       & -0.4287                        &  0.3048                        &   -0.9688                     &   0.1158                       & -0.2835                       &  \textbf{0.1529}                      &    \textbf{-0.4444}                  \\
                      & IDE                     &   0.1555                       & -0.4093                        &    0.2945                      &  -0.9115                       &  \textbf{0.1209}                        &  -0.3515                     & 0.1057                      &  -0.2831                     \\
\hline
\end{tabular}
}
\caption{Evaluation performance of feature interaction explanation methods on SST-2 and Yelp-2 datasets.}
\label{t1}

\end{table*}

For \textbf{Negative} prediction, we take two pairs $\left [ ``\mathrm{not}", ``\mathrm{might}" \right ]$ and $\left [ ``\mathrm{not}", ``\mathrm{buy}" \right ]$ as examples. In ASIV estimation, the interaction influence of $``\mathrm{might}"\rightarrow ``\mathrm{not}" $ is positively stronger than that of $``\mathrm{not}"\rightarrow``\mathrm{might}" $, while in Bivariate Shapley and Shapley interaction estimation, both asymmetric and symmetric interaction influences are negative. Intuitively, the interaction between ``not'' and ``might'' could contribute positively to \textbf{Negative} prediction. Compared with ``not'', ``might'' does not convey much information against \textbf{Negative}. Therefore $\phi_{\mathrm{not}}(\mathrm{might})<\phi_{\mathrm{might}}(\mathrm{not})$. Similarly, the interaction influence between the pair $\left [ ``\mathrm{not}", ``\mathrm{buy}" \right ]$ is positive in ASIV while in Bivariate Shapley and Shapley interaction estimation the interaction influence is negative. In practice, human evaluation tends to attribute more importance to the pair $\left [ ``\mathrm{not}", ``\mathrm{buy}" \right ]$ for model prediction. ``not'' further modifies ``buy'' so in ASIV the interaction influence of $``\mathrm{not}"\rightarrow ``\mathrm{buy}" $ could be positive and stronger than that of $``\mathrm{buy}"\rightarrow``\mathrm{not}" $.

\subsection{Quantitative Analysis}
We follow the prior works \citep{chen2020generating,guerreiro2021spectra} that set $k$ to $20$ in sentiment classification task. The sampling size $m$ is set to $500$. Due to increasing computational cost in computing each pairwise interaction (statistics of the datasets in Appendix \ref{appendixb}), we randomly choose $1000$ samples from SST-2 dataset and $100$ samples from Yelp-2 dataset that the review length is restricted to be less than $100$ words. Further, to estimate conditional expectation and in-domain expectation, in each permutation, for each token in $\bm{x}'_{\bar{S}}$ only the most likely word is sampled from the pretrained language model given $\bm{x}_{S}$.

\begin{figure*}[]
\centering
\includegraphics[width=1\linewidth,height = 4in ]{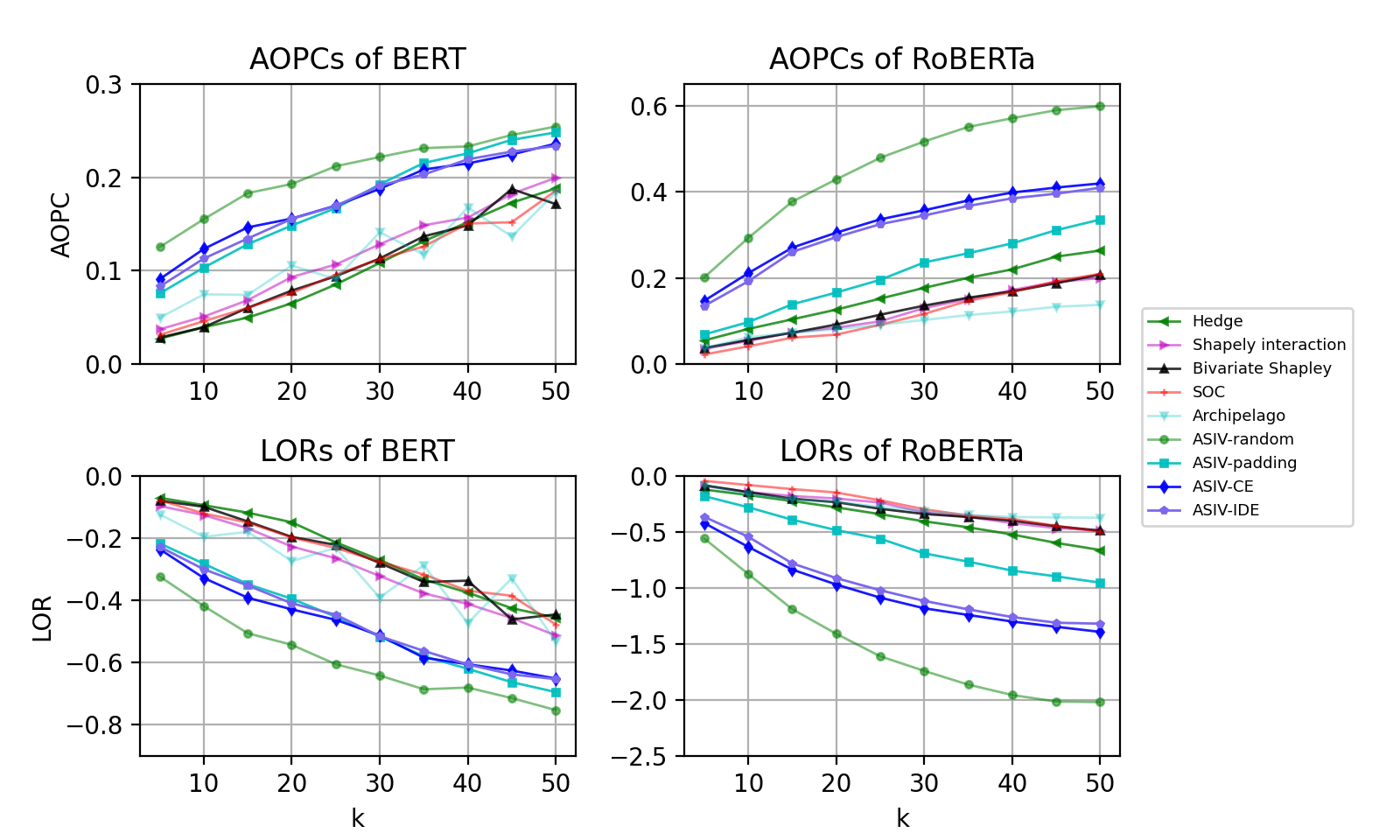}
\caption{Evaluation performance of BERT and RoBERTa on SST-2 dataset.}
\label{fig4}
\end{figure*}

\subsubsection{Comparison with baselines}
Table~\ref{t1} shows the evaluation performance of BERT and RoBERTa models on two different datasets. We observe that the proposed ASIV consistently outperforms the compared baselines in identifying influential features for the predictions of BERT and RoBERTa. ASIV with random sampling strategy performs better in SST-2 while ASIV with in-domain sampling demonstrates its effectiveness for Yelp-2 classification with RoBERTa.

We first analyze the baseline as follows: for the univariate feature attribution methods (i.e. Hedge and SOC), the explanation heavily depends on their attribution estimation. For example, compared with Shapley values, SOC could assign different values to the equivalent important features, therefore the corresponding estimated feature ranking may not be faithful. For interaction attribution methods, derived undirected interaction graph (i.e. from Archipelago and Shapley interaction index) does not emphasize the importance of some nodes in symmetric interaction. As shown in Fig.~\ref{fig11} (c), the importance of ``not'' is ignored in the interaction with ``might''. Bivariate Shapley defines a directed interaction graph but its formulation introduces noisy interaction relationship, which may result in false estimation. For example, both the directional interaction between ``not'' and ``might'' are negative in Fig.~\ref{fig11} (b). So with PageRank algorithm the obtained node importance ranking could be misleading.

For ASIV evaluated in SST-2 dataset, conditional expectation based estimation performs better than in-domain expectation and marginal expectation with padding operation while it is inferior to marginal expectation with random sampling. Since the review length in SST-2 is relatively short (see Appendix \ref{appendixb}), compared with conditional sampling, random sampling also can generate smooth and in-domain-like text sequence. For example, given the test instance in SST-2 ``\emph{the cast is uniformly excellent and relaxed}'' and the corresponding $\bm{x}_S$ ``\emph{the [MASK] is uniformly excellent [MASK] [MASK] }'', conditional sampling generates the most likely sequence ``\emph{the interior is uniformly excellent throughout .}'' and random sampling produces ``\emph{the movie is uniformly excellent and predictable}''.

In Yelp-2 dataset with RoBERTa classification, both ASIV-CE and ASIV-IDE performs better than ASIV with random sampling . Different from SST-2 reviews, the review length in Yelp-2 is quite long (see Appendix \ref{appendixb}). Random sampling is more likely to produce long disorganized $\bm{x}'_{\bar{S}}$ in some permutations. Compared with in-domain sampling where the quality of pretrained language model is restricted by the limited training corpus, conditional sampling could generate more smooth text sequences that modify the main idea of the review.

\begin{figure*}[h]
\centering
\includegraphics[width=1\linewidth,height = 3.7in ]{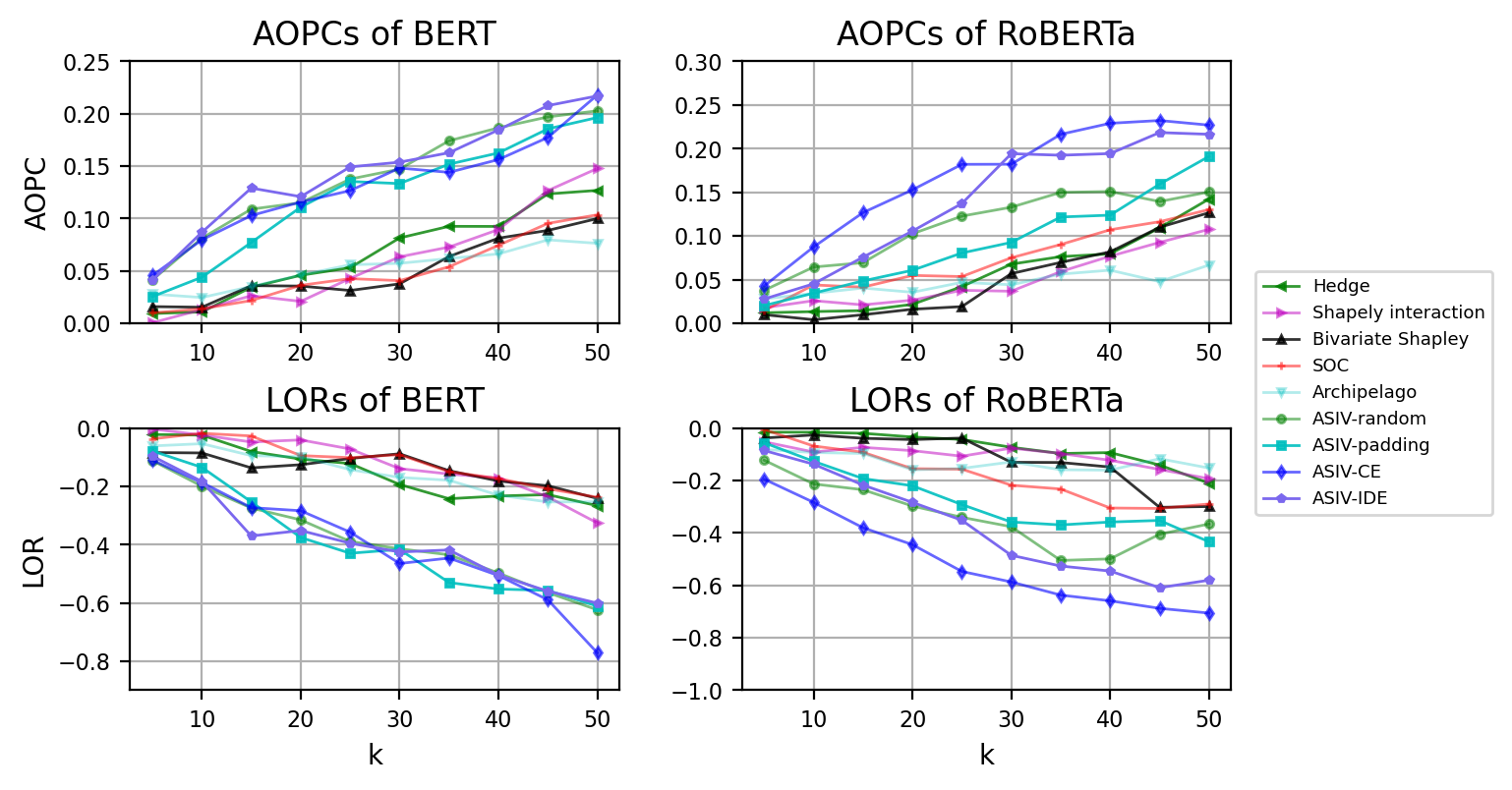}
\caption{Evaluation performance of BERT and RoBERTa on Yelp-2 dataset.} 
\label{fig5}
\end{figure*}

\subsubsection{Sensitivity analysis}
As shown in Fig.~\ref{fig4} and Fig.~\ref{fig5}, we study the influence of $k$ on the evaluation performance of BERT and RoBERTa models on SST-2 and Yelp-2 datasets.

We can see from Fig.~\ref{fig4} that ASIV always outperforms the baselines and ASIV with random sampling consistently achieves the best performance in AOPC and LOR metrics by varying $k$. With the increase of $k$, the curves of ASIV-IDE and ASIV-CE tend to overlap in BERT-based classification and ASIV-CE outperforms ASIV-IDE by a narrow margin in RoBERTa-based classification. In Yelp-2 dataset with the increase of $k$ generally both conditional and in-domain sampling strategies are more effective than random and padding operation. ASIV-IDE with BERT model performs better in AOPCs and LORs in the scenario $k<30$ and ASIV-CE with RoBERTa model performs better with $k<30$.

As addressed in the above subsection that in SST-2 the average number of words of the review is limited, random sampling could enforce the classification model to focus on the specific text span that is essential for predicting the short text sequence, while the smooth context produced by conditional and in-domain sampling strategies may contain confusing information for prediction. For predicting long text sequence, the classification model has to rely on the whole context to capture the main idea of the text. Therefore in computing marginal contribution both conditional and in-domain sampling could be more applicable for long text sequences.

\section{Conclusion}
In this paper we propose an asymmetric feature interaction attribution method to explain asymmetric higher-order feature interactions in the inference of deep neural NLP models. We extend Shapley value to asymmetric Shapley interaction value and investigate three different sampling strategies in computing marginal contribution of value function in NLP field. By evaluating our proposed model and five model-agnostic feature interaction attribution methods on two sentiment datasets with BERT and RoBERTa, our model achieves the best performance in identifying the influential words for model prediction. Also, for the three different sampling strategies we empirically show that none is generally better than the others or more broadly applicable, which could provide guidelines for the selection of reference distribution in NLP field. For example, random sampling strategy is effective for the short text sequence and for the long text sequence in-domain sampling could produce more smooth and domain-dependent context. In the future, we consider generating sparse and differential causal structure for explaining model prediction.

\section{Limitations}
The proposed asymmetric Shapley interaction value could estimate asymmetric feature interaction in explaining the prediction of deep models. There are two major concerns regarding the time complexity: estimation of marginal contribution and construction of hypergraphs. In computing value function we have to consider more permutations to reduce approximation errors. Also, before estimating the contribution of asymmetric interaction, interaction graph with different orders should be constructed. We could resort to effective approximation methods in computing marginal contribution and prior knowledge in building hypergraph.

\section*{Ethics Statement}

The authors declare that they have no conflicts of interest. This paper does not contain any studies involving business data and personal information.

\bibliography{ref}
\bibliographystyle{acl_natbib}

\appendix
\section{Implementation details of explanation methods}
\label{appendixa}

HEDGE \citep{chen2020generating}: We follow the original implementation that selects word-level features from the bottom of the hierarchical structure and obtain features' ranking based on the estimated importance. 

SOC \citep{jin2019towards}: We rank the importance of features in the bottom level of a hierarchical explanation.

Archipelago \citep{tsang2020does}: We use ArchAttribute to compute the pairwise interaction attribution and then apply PageRank algorithm.

Bivariate Shapley value \citep{masoomi2021explanations}: We follow the original paper and use Shapley sampling approximation to compute bivariate Shapley value. The sample size is set to 1000 for each interaction. 

Shapley interaction index \citep{grabisch1997k}: To ensure consistency among Shapley-based methods, we also use Shapley sampling approximation. The sample size is set to 1000 for each interaction. 

Asymmetric Shapley interaction value: For the classification with BERT model, in computing conditional expectation we employ pre-trained BERT-base model, and for in-domain expectation, we pretrain BERT-base model using training data. For RoBERTa classification model, we employ pre-trained RoBERTa-base model in estimating conditional expectation and pretrain RoBERTa-base model for in-domain expectation.

We do not select Shapley-Taylor indices for the following reason: the second-order Shapley-Taylor interaction indices for a pair $(i,j)$ with a fixed permutation $\pi$ is defined as
\begin{equation}
I_{ij,\pi}=v(S\cup ij) - v(S\cup i) - v(S\cup j) + v(S),
\end{equation}
then the second-order Shapley-Taylor interaction indices could be obtained by sampling over permutations, which is same to the computation of Shapley interaction index in our paper.

\section{Statistics of SST-2 and Yelp-2 datasets}
\label{appendixb}

Here we present the statistics of the test datasets. In Yelp-2 test set the the average review length $136.49$, to reduce computational cost in estimating pariwise interaction, we have to sample from the test with restriction to review length (i.e. review length $\leq 100$).

\begin{figure}[H]
\centering
\includegraphics[width=1\linewidth,height = 1.8in ]{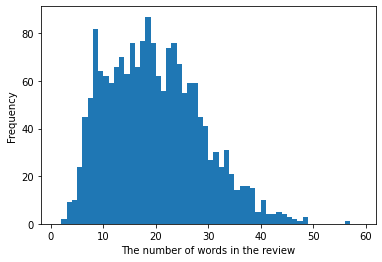}
\caption {Review length in SST-2 test set.} 
\label{1}
\end{figure}

\begin{figure}
\centering
\includegraphics[width=1\linewidth,height = 1.8in ]{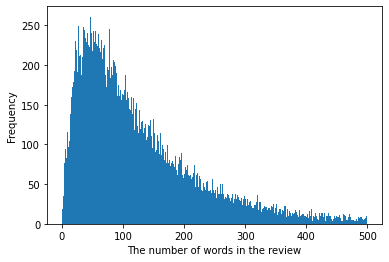}
\caption {Review length in Yelp-2 test set.} 
\label{1}
\end{figure}

\section{Examples of the estimated feature interaction graph with matrix demonstration}
\label{appendixc}

\begin{figure}[H]
\centering
\begin{subfigure}[b]{0.55\textwidth}
   \includegraphics[width=0.9\linewidth,height=4cm]{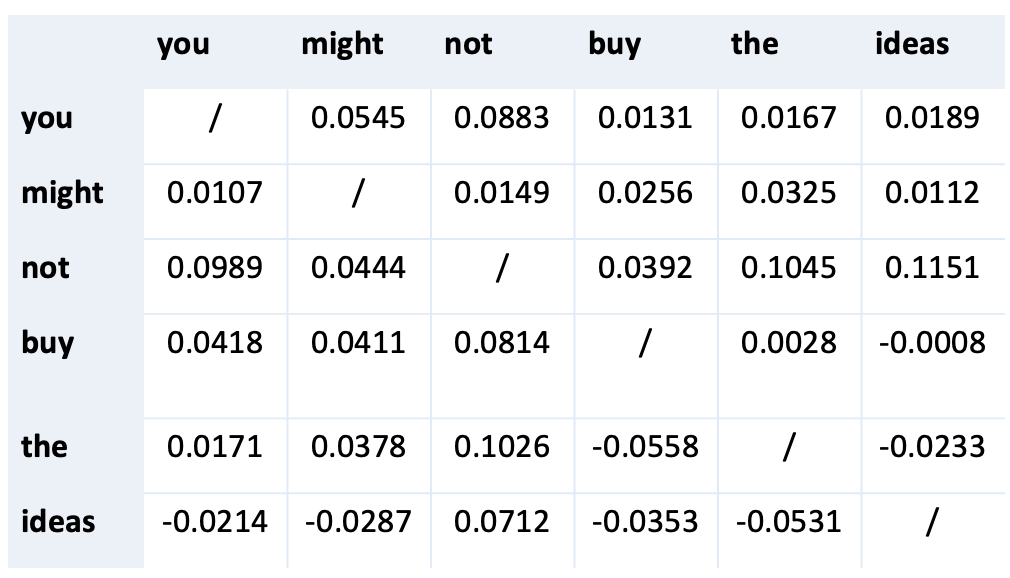}
   \caption{ASIV estimation.}
   \label{fig:Ng1} 
\end{subfigure}

\begin{subfigure}[b]{0.55\textwidth}
   \includegraphics[width=0.9\linewidth,height=4cm]{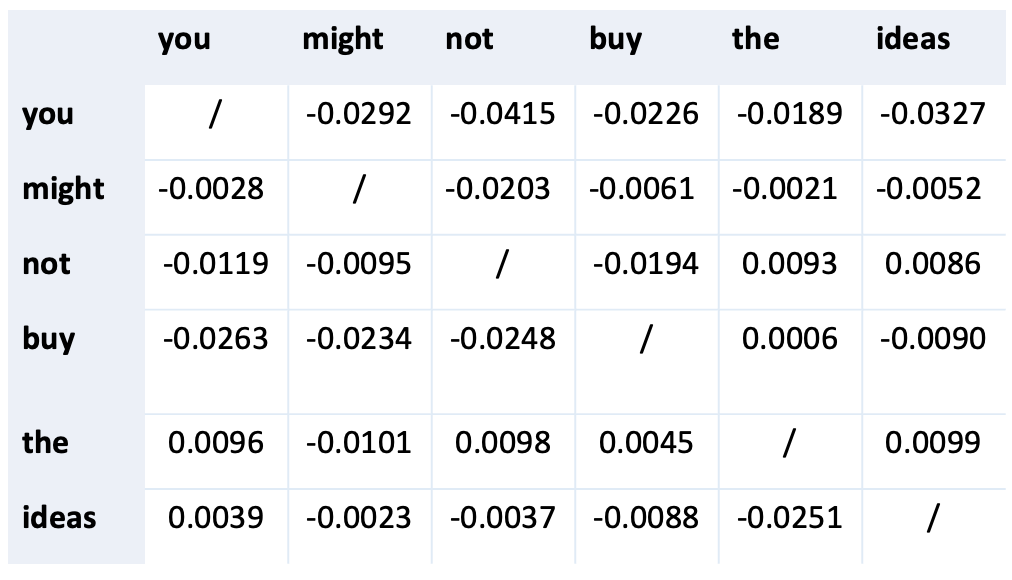}
   \caption{Bivariate Shapley estimation.}
   \label{fig:Ng2}

\end{subfigure}

 \begin{subfigure}[b]{0.55\textwidth}
   \includegraphics[width=0.9\linewidth,height=4cm]{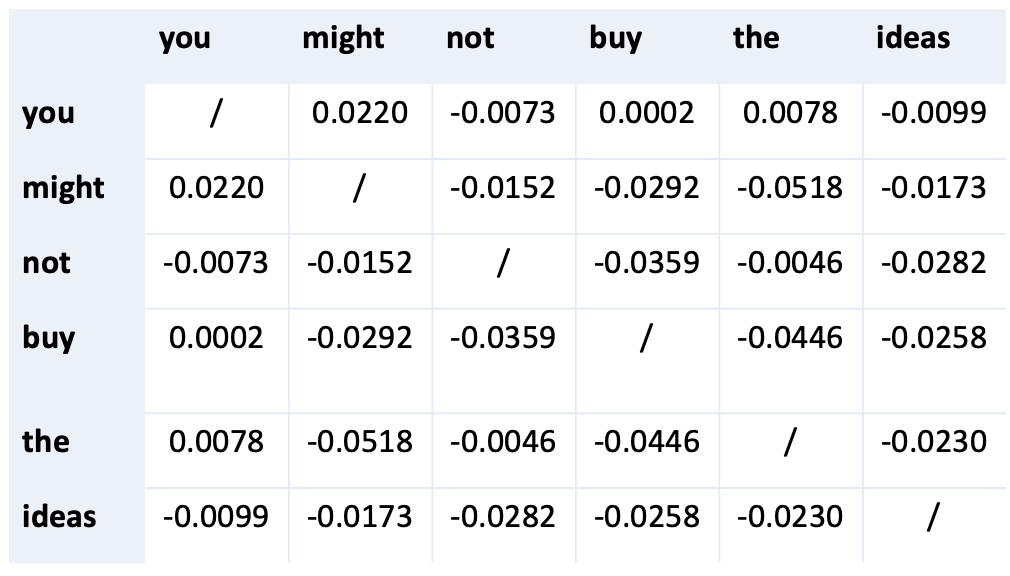}
   \caption{Shapley interaction index estimation.}
   \label{fig:Ng2}  
\end{subfigure}

\caption{Matrix of the estimated feature interaction graph for the instance ``you might not buy the ideas'' from SST-2 that is predicted as \textbf{Negative} by BERT model.}
\end{figure}

\makeatletter
\setlength{\@fptop}{0pt}
\makeatother
\begin{figure}[ht!]
\centering
\begin{subfigure}[b]{0.55\textwidth}
   \includegraphics[width=0.8\linewidth,height=4cm]{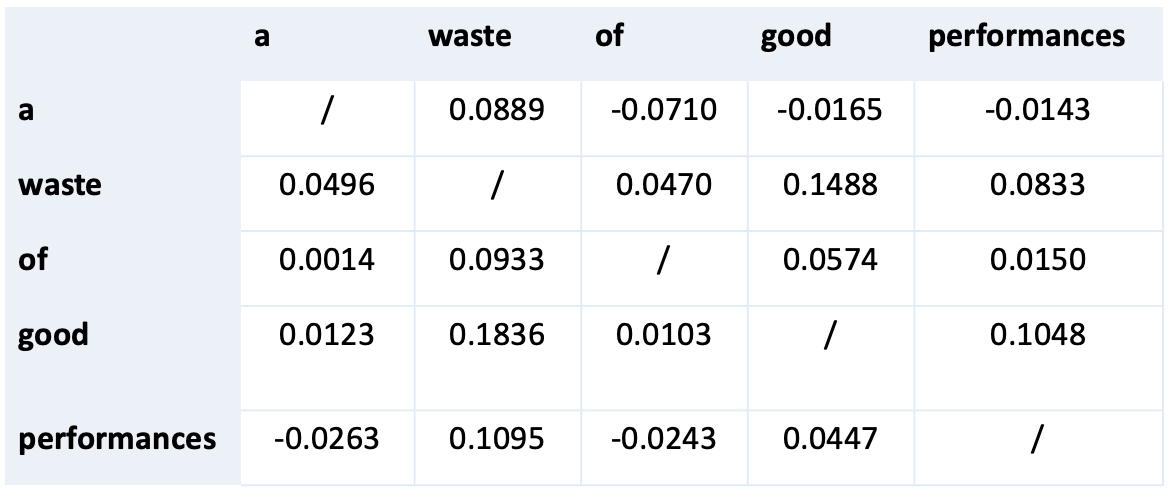}
   \caption{ASIV estimation.}
   \label{fig:Ng1} 
\end{subfigure}

\begin{subfigure}[b]{0.55\textwidth}
   \includegraphics[width=0.8\linewidth,height=4cm]{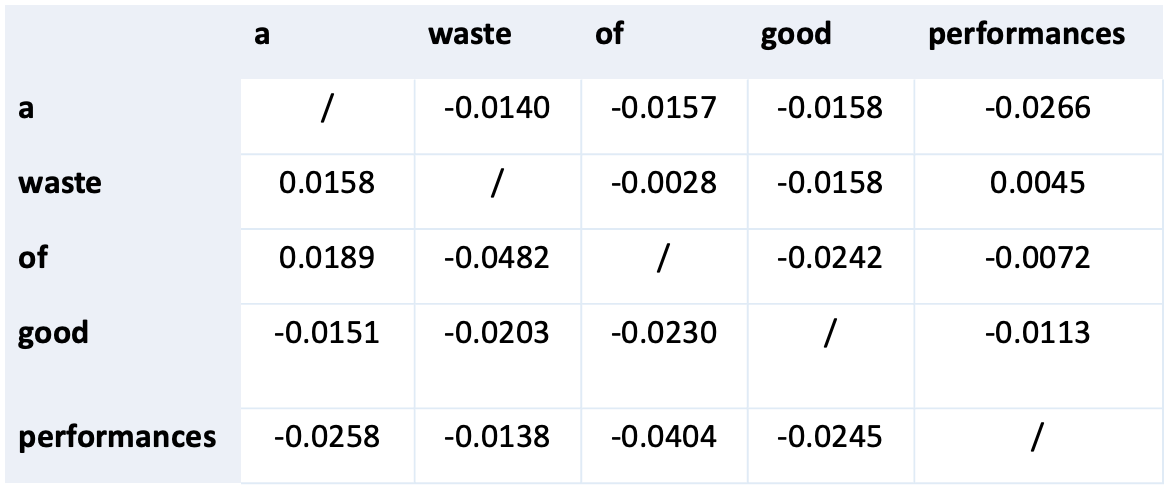}
   \caption{Bivariate Shapley estimation.}
   \label{fig:Ng2}

\end{subfigure}

 \begin{subfigure}[b]{0.55\textwidth}
   \includegraphics[width=0.8\linewidth,height=4cm]{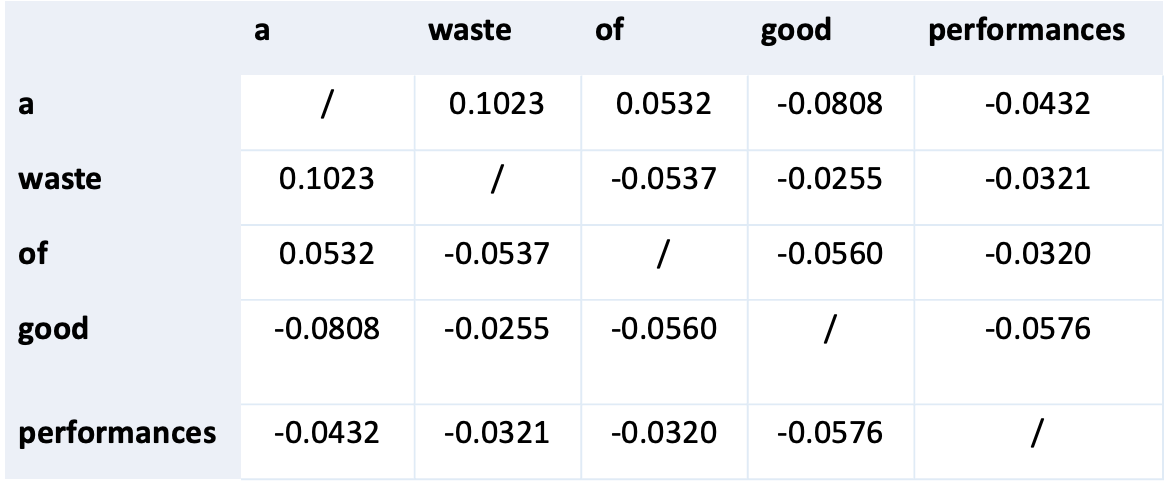}
   \caption{Shapley interaction index estimation.}
   \label{fig:Ng2}  
\end{subfigure}

\caption{Matrix of the estimated feature interaction graph for the instance ``a waste of good performance'' from SST-2 that is predicted as \textbf{Negative} by BERT model.}
\end{figure}

\end{document}